\documentclass[lettersize,journal]{IEEEtran}
\usepackage{amsmath,amssymb,amsfonts}
\usepackage{algorithmic}
\usepackage{algorithm}
\usepackage{array}
\usepackage{textcomp}
\usepackage{url}
\usepackage{verbatim}
\usepackage{subfigure}
\usepackage{graphicx}
\usepackage{cite}
\usepackage{booktabs}
\usepackage{multirow}
\usepackage{makecell}

\usepackage{hyperref}
\usepackage{bm}
\usepackage{}
\usepackage{comment}

\newcommand{\etal}{\textit{et al}. }

\bstctlcite{IEEEexample:BSTcontrol}

\begin{document}

\title{4D-CAAL: \underline{4D} Radar-Camera \underline{Ca}libration and \underline{A}uto-\underline{L}abeling for Autonomous Driving}

\author{
Shanliang Yao,
Zhuoxiao Li, 
Runwei Guan, 
Kebin Cao,
Meng Xia,
Fuping Hu,
Sen Xu,
Yong Yue, 
Xiaohui Zhu,
Weiping Ding,
Ryan Wen Liu

\thanks{$^{\text{1}}$ Shanliang Yao is with the School of Information Engineering, Yancheng Institute of Technology, Yancheng 224051, China, School of Navigation, Wuhan University of Technology, Wuhan 430063, China, and also with Hubei Key Laboratory of Inland Shipping Technology, Wuhan 430063, China. (email: shanliang.yao@ycit.edu.cn).}
\thanks{$^{\text{2}}$ Zhuoxiao Li and Runwei Guan are with The Hong Kong University of Science and Technology (Guangzhou), Guangzhou 511400, China. (email: \{zhuoxiaoli, runweiguan\}@hkust-gz.edu.cn).}
\thanks{$^{\text{3}}$ Kebin Cao, Meng Xia, Fuping Hu and Sen Xu are with the School of Information Engineering, Yancheng Institute Technology, Yancheng 224051, China. (email: ckb0256@stu.ycit.edu.cn, \{shom, hufuping\}@ycit.edu.cn, xusen@ycit.cn).}
\thanks{$^{4}$ Yong Yue and Xiaohui Zhu are with School of Advanced Technology, Xi'an Jiaotong-Liverpool University, Suzhou 215123, China. (email: \{yong.yue, xiaohui.zhu\}@xjtlu.edu.cn).}
\thanks{$^{5}$ Weiping Ding is with School of Information Science and Technology, Nantong University, Nantong 226019, China. (email: dwp9988@163.com).}
\thanks{$^{6}$ Ryan Wen Liu is with School of Navigation, Wuhan University of Technology, Wuhan 430063, China, and also with the State Key Laboratory of Maritime Technology and Safety, Wuhan 430063, China. (email: wenliu@whut.edu.cn).}
}

\maketitle

\begin{abstract}
4D radar has emerged as a critical sensor for autonomous driving, primarily due to its enhanced capabilities in elevation measurement and higher resolution compared to traditional 3D radar.
Effective integration of 4D radar with cameras requires accurate extrinsic calibration, and the development of radar-based perception algorithms demands large-scale annotated datasets. However, existing calibration methods often employ separate targets optimized for either visual or radar modalities, complicating correspondence establishment. Furthermore, manually labeling sparse radar data is labor-intensive and unreliable. 
To address these challenges, we propose 4D-CAAL, a unified framework for 4D radar-camera calibration and auto-labeling. Our approach introduces a novel dual-purpose calibration target design, integrating a checkerboard pattern on the front surface for camera detection and a corner reflector at the center of the back surface for radar detection. We develop a robust correspondence matching algorithm that aligns the checkerboard center with the strongest radar reflection point, enabling accurate extrinsic calibration. Subsequently, we present an auto-labeling pipeline that leverages the calibrated sensor relationship to transfer annotations from camera-based segmentations to radar point clouds through geometric projection and multi-feature optimization.
Extensive experiments demonstrate that our method achieves high calibration accuracy while significantly reducing manual annotation effort, thereby accelerating the development of robust multi-modal perception systems for autonomous driving.
\end{abstract}


\vspace{\baselineskip}

\section{Introduction}
Radar sensors are essential for autonomous driving due to their robust all-weather operation and long-range detection capabilities \cite{yao2025exploring}. The transition from 3D to 4D imaging radar has further enhanced these systems by providing elevation information in addition to range, azimuth, and velocity measurements, which significantly improves spatial resolution and environmental perception \cite{han20234d, yao2025usvtrack}. The direct Doppler velocity measurement of 4D radar offers critical motion cues for dynamic object tracking. Additionally, its cost-effectiveness and reliability at distances exceeding 400 meters make it suitable for commercial deployment \cite{pan2024ratrack, zheng2022tj4dradset}. As autonomous vehicles progress toward higher levels of automation, integrating 4D radar with cameras has become essential for comprehensive multi-modal perception \cite{yao2023radar, yao2024waterscenes, an2025embodied}.


Despite these advantages, 4D radar encounters significant challenges, such as sparse point clouds, multi-path reflections, and ghost targets \cite{engels2021automotive}. Limited angular resolution results in imprecise localization, especially for small or distant objects. Addressing these limitations requires accurate extrinsic calibration between 4D radar and cameras to enable effective sensor fusion. This integration allows systems to combine radar's velocity measurements and all-weather reliability with the camera's rich semantic information \cite{park2014calibration}. Additionally, deep learning-based radar perception relies on large-scale annotated datasets. Manual annotation of sparse and ambiguous radar data is labor-intensive and prone to errors \cite{nobis2019deep}, highlighting the need for automated labeling methods to advance radar-based perception.

While extensive research exists on LiDAR-camera calibration and auto-labeling \cite{cui2021deep, pandey2015automatic}, the integration of 4D radar and camera systems remains insufficiently explored. Current radar-camera calibration methods typically use separate targets optimized for either visual or radar modalities, necessitating multiple calibration objects or complex experimental setups \cite{barjenbruch2015joint}. Furthermore, LiDAR-based auto-labeling pipelines are not directly applicable to 4D radar because of fundamental differences in sensing principles and data characteristics \cite{zhang2021survey}. The absence of integrated frameworks that connect calibration with subsequent auto-labeling processes restricts the efficiency and scalability of preparing 4D radar datasets for perception algorithm development.

In response to these challenges, we propose 4D-CAAL, a unified framework for 4D radar-camera calibration and auto-labeling. Our approach introduces a novel dual-purpose calibration target design that integrates complementary features observable by both sensing modalities. Specifically, we develop an A4-sized calibration board with a checkerboard pattern on the front surface for camera-based corner detection and a trihedral corner reflector mounted at the center of the back surface for radar-based localization. Moreover, our calibration methodology leverages robust correspondence matching by aligning the geometric center of the checkerboard pattern extracted from camera images with the centroid of the strongest radar reflection point from the corner reflector. 

Once accurate calibration is achieved, we employ this spatial relationship to enable efficient automatic annotation of radar point clouds. Our auto-labeling pipeline transfers high-quality segmentation results from the camera domain to the radar point cloud through geometric projection and spatial association. The framework then applies temporal consistency filtering and spatial clustering techniques to radar features to refine the projected labels, removing erroneous associations and enhancing annotation quality. 
Extensive experiments conducted in real-world autonomous driving scenarios demonstrate that the proposed method achieves accurate calibration. The automatically generated labels exhibit high consistency with manual annotations and support the training of radar-based detection models with competitive performance.
The key contributions of our study are summarized as follows:
\begin{itemize}
\item We design a compact dual-purpose calibration target that combines a checkerboard pattern with a central corner reflector on opposite sides of an A4-sized board, enabling simultaneous feature extraction and correspondence establishment for both 4D radar and camera modalities.
\item We propose a robust radar-camera calibration algorithm that establishes geometric correspondences by aligning the checkerboard center with the strongest radar reflection point, achieving accurate extrinsic parameter estimation.
\item We develop a coarse-to-fine auto-labeling framework that effectively combines visual semantic understanding with 4D radar spatial and motion features.
\item We conduct comprehensive experiments in real-world autonomous driving scenarios, demonstrating that our framework achieves calibration accuracy comparable to existing methods while enabling efficient generation of high-quality radar annotations.
\end{itemize}

The remainder of this paper is structured as follows. Section \ref{sec:related-work} reviews target-based calibration and point‑based auto‑labeling. Section \ref{sec:methodology} introduces the proposed 4D‑CAAL framework, which comprises geometric correspondence calibration and a coarse‑to‑fine auto‑labeling pipeline. Section \ref{sec:experiments} presents experimental results, comparative analysis, and discussion. Finally, Section \ref{sec:conclusion} concludes the study and outlines future research directions.

\section{Related Work}
\label{sec:related-work}
\subsection{Target-Based Radar-Camera Calibration}
Target-based calibration methods are considered the ideal approach for radar-camera extrinsic calibration due to their superior accuracy and reliability. These methods rely on well-designed calibration targets that provide precise, simultaneously observable pose features for both sensor modalities.

\subsubsection{Integrated Calibration Board Designs}
To address the challenge of simultaneous visibility across heterogeneous sensors, researchers have developed integrated calibration targets that combine visual and radar-reflective features. Per{\v{s}}i{'c} \etal introduced a dual-modality calibration board consisting of a styrofoam triangle with a checkerboard pattern on one surface and a triangular corner reflector on the opposite side \cite{pervsic2019extrinsic}. Their method employs a two-step optimization approach that minimizes both reprojection errors and radar cross-section (RCS) inconsistencies, achieving reliable calibration across multiple radar models. Building upon this concept, Domhof \etal proposed an advanced calibration target featuring a 1.0m × 1.5m styrofoam board with four circular holes and a trihedral corner reflector positioned at the center \cite{domhof2019extrinsic}. The integration of multiple geometric features, such as checkerboard corners, circular holes, and corner reflectors, enables the establishment of correspondences across three sensing modalities simultaneously.

\subsubsection{4D Radar-Specific Calibration}
The advent of 4D imaging radar with elevation resolution has motivated specialized calibration methodologies that exploit the unique characteristics of 4D radar data. Cui \etal presented one of the first target-based calibration methods specifically designed for 4D radar-camera systems, addressing the additional elevation dimension and its associated challenges \cite{cui20213d}. More recently, researchers have developed uncertainty-aware approaches that explicitly model the spherical coordinate noise propagation inherent in radar measurements. These methods account for the fact that radar detection uncertainties exhibit non-uniform distributions in Cartesian space due to the nonlinear transformation from spherical coordinates (range, azimuth, elevation). By incorporating 3D uncertainty models into the perspective-n-point (PnP) optimization process, such approaches achieve improved robustness against sensor noise and environmental variability \cite{cao20254d}.

\subsection{Point-Based Auto-Labeling}
Data labeling for automotive radar point clouds is a highly labor-intensive and time-consuming task. Unlike LiDAR, radar point clouds are inherently sparse and ambiguous, making it nearly impossible to discern the physical shapes of objects directly from raw radar returns \cite{yao2023radar}.

\subsubsection{Multi-modal Assisted Auto-Labeling} 
Leveraging complementary sensing modalities has emerged as the dominant paradigm for automated radar labeling. 
Camera-aided methods transfer visual detections to radar through sensor fusion and correspondence matching. Sengupta \etal pioneered an influential framework combining a pre-trained YOLOv3 detector with the Hungarian algorithm for optimal radar-image association \cite{sengupta2022automatic}. LiDAR-assisted methods offer superior geometric fidelity for radar annotation. Cheng \etal proposed annotating radar data using synchronized LiDAR point clouds via 3D Euclidean distance-based association \cite{cheng2021new}. Inspired by this research, Sun \etal recently introduced an efficient 4D radar tensor auto-labeling method that trains a LiDAR-based object detection network on high-quality calibrated LiDAR data, which then autonomously generates radar labels without human intervention \cite{sun2024efficient}. 

Semi-automatic frameworks balance automation efficiency with quality assurance. The Astyx dataset employs automatic pre-labeling with Active Learning-based uncertainty sampling to selectively refine labels requiring manual correction \cite{meyer2019automotive}. Dimitrievski \etal developed a semi-automatic tool that generates initial annotations using state-of-the-art recognition algorithms with minimal manual intervention \cite{dimitrievski2020weakly}. 
   
\subsubsection{Label Quality Enhancement and Clutter Filtering} 
A critical challenge in radar auto-labeling is the prevalence of clutter, which introduces significant noise into labeled datasets. Because radar point clouds inherently contain scattered clutter around genuine targets, specialized filtering methods are essential to improve label quality. Cheng \etal introduced the Radar Point Cloud Spatiotemporal Filter (RSTF) that removes clutter by analyzing both spatial neighborhoods and temporal consistency \cite{cheng2022novel}. The spatial filter compares the motion-compensated Doppler velocity of each point with statistical thresholds computed for its grid cell, while the temporal filter eliminates outliers that lack stability across frames. This two-stage approach significantly enhances the signal-to-clutter ratio. 

Complementing filtering techniques, automated clutter-annotation methods have been proposed to generate ground-truth clutter labels for training dedicated clutter-detection models \cite{kopp2023tackling}. These approaches employ neural network architectures such as PointNet++ \cite{qi2017pointnet++} with customized input representations to identify clutter in multi-scan radar point clouds. By creating publicly available clutter datasets and demonstrating superior detection performance, these methods provide crucial resources to ensure that automatically generated radar labels maintain sufficient quality for training robust perception models, as unfiltered, noisy labels can severely degrade model performance and generalization.

\section{Methodology}
\label{sec:methodology}
\subsection{4D-CAAL Pipeline}
We propose 4D-CAAL, a two-stage framework for 4D radar-camera calibration and automated label generation. In the first stage, we calibrate the extrinsic parameters using a custom target that combines a corner reflector and a checkerboard, optimizing the transformation by minimizing reprojection errors across multiple poses. 
In the second stage, we leverage the calibrated parameters to generate point-level labels for radar data automatically. 
This integrated pipeline enables accurate sensor fusion and scalable label generation for 4D radar perception systems.

\subsection{\underline{4D} Radar-Camera \underline{Ca}libration}
We present a 4D radar-camera extrinsic calibration framework to estimate the rigid transformation between the sensor coordinate systems. Our approach has five main parts: 1) a calibration platform with a custom target that combines a trihedral corner reflector and a checkerboard pattern, mounted with a rigid bracket for stable alignment; 2) collecting data from multiple poses to provide enough geometric constraints; 3) extracting camera features to find the checkerboard center in the image; 4) extracting radar features by filtering and clustering the 4D radar point cloud to detect the corner reflector; and 5) optimizing the extrinsic parameters by minimizing the reprojection error between matching 2D and 3D points. The following subsections explain each part in detail.

\subsubsection{Calibration Platform}
\paragraph{Calibration Board Design}
\begin{figure}
\centering
\includegraphics[width=1\linewidth]{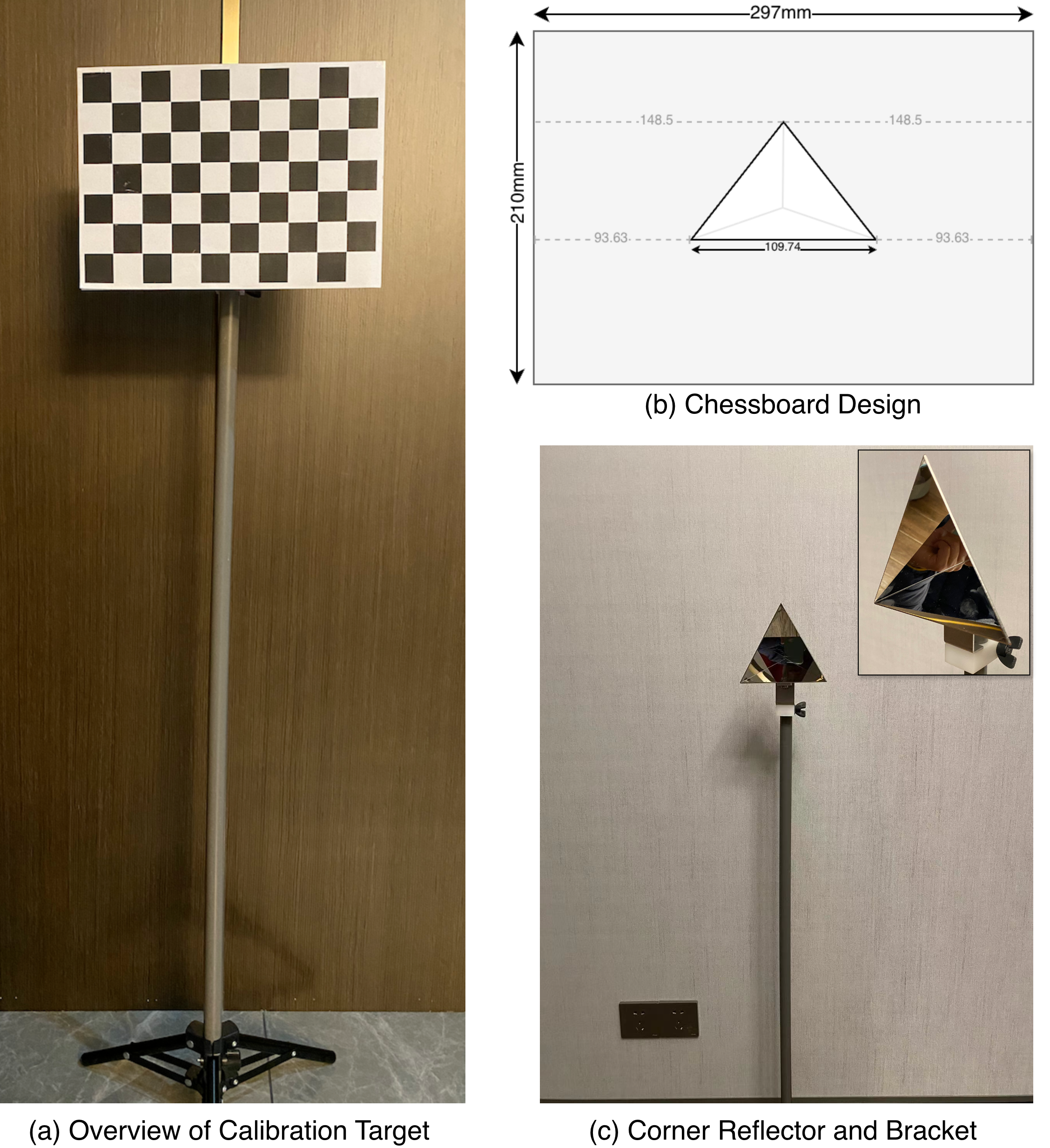} 
\caption{Calibration board design. Our custom calibration target integrates a trihedral corner reflector with an A4-sized acrylic board featuring a checkerboard pattern. The camera is calibrated by associating the detected checkerboard center with the known 3D coordinates of the corner reflector.}
\label{fig:calibration-board}
\end{figure}

A well-designed calibration target is key to accurate extrinsic calibration. The target must meet two main criteria: it should be easy for both radar and camera to detect, and it should allow for precise localization in both systems. To achieve this, we created a new calibration target that combines a trihedral corner reflector with an acrylic board featuring a checkerboard pattern, as shown in Figure \ref{fig:calibration-board}. The checkerboard’s black-and-white squares make it easy for the camera to detect location features. The acrylic board is standard A4 size (210 mm $\times$ 297 mm), with a metal reflector mounted at the center on the back. We calibrate the camera’s extrinsic parameters, including the rotation matrix and translation vector, by detecting the checkerboard center and matching it to the known 3D coordinates of the reflector.

\paragraph{Rigid Mounting Bracket Design}
\begin{figure}
\centering
\includegraphics[width=1\linewidth]{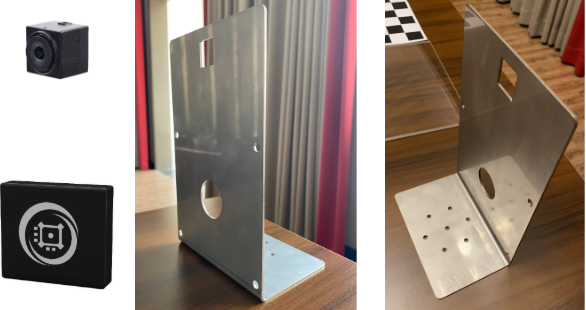} 
\caption{Rigid Mounting Bracket. The high‑strength steel bracket ensures precise sensor alignment and includes standardized lower holes for vehicle integration.}
\label{fig:fixed}
\end{figure}

To keep the 4D radar and camera in stable positions, we designed a custom rigid mounting bracket made from high-strength steel, as shown in Figure \ref{fig:fixed}. The bracket aligns the sensor mounting positions with the mounting holes of both sensors.
Additionally, to facilitate integration with various autonomous vehicle platforms, the bracket incorporates standardized mounting holes along its lower section. The robust steel structure provides excellent mechanical stability under operational conditions, effectively mitigating vibrations and resisting environmental factors such as temperature variations and weather exposure. 

\subsubsection{Data Collection}
For extrinsic calibration, we collected measurements of the calibration target from different positions and angles. This provided enough information and constraints for the optimization process that estimates the relative pose between the radar and camera.
Through detection on the calibration target, $K$ pairs of corresponding measurements are obtained from the two sensors. The camera provides RGB images, and the radar outputs point clouds containing range, azimuth, elevation, radial velocity, and reflectivity.

\subsubsection{Camera Feature Extraction}
For each calibration pose, RGB images are captured by the camera. The pixel coordinates of the checkerboard corners are extracted using standard corner detection and subsequent sub-pixel refinement. The complete set of detected corners is defined in Equation \ref{eq:camera-feature-extraction}:
\begin{equation}
\label{eq:camera-feature-extraction}
\mathcal{C} = \{c_1, c_2, \ldots, c_n\}, \quad \mathbf{c}_i = (u_i, v_i) \in \mathbb{R}^2
\end{equation}
where $n = (N_x - 1) \times (N_y - 1)$, where $N_x$ and $N_y$ represent the number of checkerboard squares along the $x$ and $y$ axes, respectively.

The geometric centroid of all detected corners is computed to determine the center of the reflector in the image plane, as defined in Equation \ref{eq:camera-center}:
\begin{equation}
\label{eq:camera-center}
\mathbf{c}_{\text{center}} = \frac{1}{n} \sum_{i=1}^n \mathbf{c}_i = \left( \frac{1}{n} \sum_{i=1}^n u_i, \frac{1}{n} \sum_{i=1}^n v_i \right)
\end{equation}

The corresponding pseudo-code for camera feature extraction is summarized in Algorithm \ref{alg:code-camera}.

\begin{algorithm}[t]
\caption{Camera Feature Extraction}
\label{alg:code-camera}
\begin{algorithmic}[1]
\REQUIRE Input image $\mathbf{I}$, checkerboard grid size $(N_x, N_y)$
\ENSURE Detected corner points $\mathcal{C}$, reflector center $\mathbf{c}_{\text{center}}$
\STATE // Detect checkerboard corners
\STATE $\mathcal{C}_{\text{raw}} \leftarrow \texttt{detectChessboardCorners}(\mathbf{I}, (N_x, N_y))$
\STATE // Refine to sub-pixel accuracy
\STATE $\mathcal{C} \leftarrow \texttt{cornerSubPix}(\mathbf{I}, \mathcal{C}_{\text{raw}})$
\STATE // Compute geometric centroid as reflector center
\STATE $\mathbf{c}_{\text{center}} \leftarrow \frac{1}{|\mathcal{C}|} \sum_{\mathbf{c}_i \in \mathcal{C}} \mathbf{c}_i$
\RETURN $\mathcal{C}, \mathbf{c}_{\text{center}}$
\end{algorithmic}
\end{algorithm}

\subsubsection{Radar Corner Reflector Extraction}
The extraction of the corner reflector center from 4D radar data involves a four-step pipeline: a) transforming the coordinates, b) applying preprocessing filters, c) performing spatial clustering, and d) localizing the center point.

\paragraph{Coordinate Transformation}
The 4D radar outputs raw point cloud measurements in spherical coordinates, as shown in Equation \ref{eq:d}.
\begin{equation}
\label{eq:d}
\mathcal{D}=\left\{d_1, d_2, \ldots, d_n\right\}, \quad d_i=\left\{R_i, \theta_i, \phi_i, v_i, \rho_i \right\}
\end{equation}
where $R_i$ is range, $\theta_i$ is azimuth angle, $\phi_i$ is elevation angle, $v_i$ is radial velocity, and $\rho_i$ is radar cross-section intensity.

\paragraph{Preprocessing Filtering}
To isolate the corner reflector from environmental clutter, range, velocity, and RCS constraints are applied, as expressed in Equation \ref{eq:radar-filtering}:
\begin{equation}
\mathcal{D}_{\text{filtered}} = \left\{d_i \in \mathcal{D} \;\middle|\; 
\begin{array}{l}
R_{\min} \leq R_i \leq R_{\max}, \\
|v_i| < v_{\text{th}}, \\
\rho_i > \rho_{\min}
\end{array}
\right\}
\label{eq:radar-filtering}
\end{equation}
where $R_{\min}$ and $R_{\max}$ define the calibration working range, $v_{\text{th}}$ is the velocity threshold for static points, and $\rho_{\min}$ is the minimum intensity threshold. This step leverages the corner reflector's characteristics to distinguish it from background objects.

\paragraph{Spatial Clustering}
Density-Based Spatial Clustering of Applications with Noise (DBSCAN) is employed to group filtered points, as defined in Equation \ref{eq:dbscan}:
\begin{equation}
\{C_1, \ldots, C_m\} \gets \text{DBSCAN}(\mathcal{D}_{\text{filtered}}^{xyz}, \epsilon, N_{\min})
\label{eq:dbscan}
\end{equation}
where $\mathcal{D}{\text{filtered}}^{xyz} = {\mathbf{p}i^r \mid d_i \in \mathcal{D}{\text{filtered}}}$ is the set of filtered 3D coordinates, $\epsilon$ is the neighborhood radius, and $N{\min}$ is the minimum number of points to form a dense region.
Thus, this step identifies clusters ${C_1, C_2, \ldots, C_m}$ by connecting points within $\epsilon$-neighborhoods. Points that cannot be linked to a dense region are discarded as noise.

\paragraph{Center Point Localization}
Among all detected clusters, the corner reflector is identified as the cluster with maximum reflected intensity, as shown in Equation \ref{eq:ccorner}:
\begin{equation}
C_{\text{corner}} = \arg\max_{C_j} \frac{1}{|C_j|} \sum_{d_i \in C_j} \rho_i
\label{eq:ccorner}
\end{equation}
where $|C_j|$ denotes the cardinality of cluster $C_j$. This criterion leverages the fact that corner reflectors produce significantly stronger radar returns than environmental scatterers.

The 3D center of the corner reflector in the radar frame is determined by selecting the point with maximum RCS within the identified cluster, as defined in Equation \ref{eq:dcenter}:
\begin{equation}
\mathbf{d}_\text{center} = \arg\max_{d_i \in C_{\text{corner}}} \rho_i
\label{eq:dcenter}
\end{equation}

Overall, the pseudo-code for radar feature extraction is summarized in Algorithm \ref{alg:code-radar}.

\begin{algorithm}[t]
\caption{Radar Corner Reflector Extraction}
\label{alg:code-radar}
\begin{algorithmic}[1]
\REQUIRE Raw radar point cloud $\mathcal{D} = \{d_1, \ldots, d_n\}$, where $d_i = (R_i, \theta_i, \phi_i, v_i, \rho_i)$
\REQUIRE Filter thresholds: $R_{\min}, R_{\max}, v_{\text{th}}, \rho_{\min}$; DBSCAN parameters: $\epsilon, N_{\min}$
\ENSURE 3D position of the corner reflector $\mathbf{d}_{\text{center}}$
\STATE // Step 1. Preprocessing Filtering
\STATE $\mathcal{D}_{\text{filtered}} \leftarrow$ Apply constraints from Equation (\ref{eq:radar-filtering}) to $\mathcal{D}$
\STATE $\mathcal{D}_{\text{filtered}}^{xyz} \leftarrow \{\texttt{sph2cart}(R_k, \theta_k, \phi_k) \mid d_k \in \mathcal{D}_{\text{filtered}}\}$
\STATE // Step 2. Spatial Clustering (Equation \ref{eq:dbscan})
\STATE $\{C_1, \ldots, C_m\} \leftarrow \text{DBSCAN}(\mathcal{D}_{\text{filtered}}^{xyz}, \epsilon, N_{\min})$ 
\STATE // Step 3. Corner Reflector Identification (Equation \ref{eq:ccorner})
\STATE $C_{\text{corner}} \leftarrow \arg\max_{C_j} \frac{1}{|C_j|} \sum_{d_i \in C_j} \rho_i$
\STATE // Step 4. Center Point Localization (Equation \ref{eq:dcenter})
\STATE $\mathbf{d_{\text{corner}}} \leftarrow \arg\max_{d_i \in C_{\text{corner}}} \rho_i$
\STATE $\mathbf{d}_{\text{center}} \leftarrow \text{sph2cart}(R_{\text{center}}, \theta_{\text{center}}, \phi_{\text{center}})$
\RETURN $\mathbf{d}_{\text{center}}$
\end{algorithmic}
\end{algorithm}

\subsubsection{Extrinsic Calibration Procedure}
The extrinsic calibration procedure estimates the rigid-body transformation between the radar and camera coordinate frames. The process is achieved through four key steps: a) establishing 2D-3D data correspondences via spatio-temporal synchronization, b) defining the transformation model between coordinate frames, c) formulating the reprojection error using the pinhole camera model, and d) solving the resulting optimization problem to obtain the optimal extrinsic parameters.

\paragraph{Data Correspondence}
Through spatio-temporal synchronization, the image plane center point $\mathbf{c}_{\text{center}}$ detected by the camera is aligned and paired with the 3D spatial point $\mathbf{d}_{\text{center}}$ detected by the radar. As presented in Equation \ref{eq:dc}, these matched pairs form the correspondence set for extrinsic calibration:
\begin{equation}
\label{eq:dc}
\mathcal{D} = \left\{(\mathbf{c}_{\text{center}}^i, \mathbf{d}_{\text{center}}^i)\right\}_{i=1}^K
\end{equation}
where $K$ denotes the number of valid calibration poses, $\mathbf{c}_{\text{center}}^{i} = (u_i, v_i)^\top \in \mathbb{R}^2$ represents the 2D image coordinates, and $\mathbf{d}_{\text{center}}^{i} = (x_i, y_i, z_i)^\top \in \mathbb{R}^3$ represents the 3D point in the radar coordinate frame.

\paragraph{Transformation Model}
The extrinsic relationship between the radar and camera coordinate frames is represented by a rigid body transformation matrix $\mathbf{T} \in \text{SE}(3)$, which projects points from the radar frame to the camera frame. As defined in Equation \ref{eq:t}, the transformation matrix consists of a rotation matrix $\mathbf{R} \in \text{SO}(3)$ and a translation vector $\mathbf{t} \in \mathbb{R}^3$:
\begin{equation}
\label{eq:t}
    \mathbf{T} = 
    \begin{bmatrix}
         \mathbf{R} & \mathbf{t} \\
         \mathbf{0} & 1
    \end{bmatrix} \in \mathbb{R}^{4 \times 4}
\end{equation}

\paragraph{Projection Model}
Using the pinhole camera model, the projection function $\pi: \mathbb{R}^3 \rightarrow \mathbb{R}^2$ maps a 3D point from the camera frame onto the 2D image plane, as shown in Equation \ref{eq:pi}:
\begin{equation}
\label{eq:pi}
\pi(\mathbf{K}, \mathbf{T}, \mathbf{d})=\binom{\hat{u}}{\hat{v}}=\binom{f_x \frac{X}{Z}+c_x}{f_y \frac{Y}{Z}+c_y}
\end{equation}
where $\mathbf{K}$ is camera intrinsic matrix, $(f_x, f_y)$ denote the focal lengths and $(c_x, c_y)$ denote the principal point coordinates.

For the $i$-th measurement pair, the reprojection error is defined as the squared Euclidean distance between the observed image center $\mathbf{c}_{\text{center}}^{i}$ and the projected radar point $\pi(\mathbf{K}, \mathbf{T}, \mathbf{d}_{\text{center}}^{i})$, as defined in Equation \ref{eq:ei}:
\begin{equation}
\label{eq:ei}
\begin{aligned}
e_i&=\left\|c_{\text{center}}^{i}-\pi\left(\mathbf{K}, \mathbf{T}, d_{\text {center}}^{i}\right)\right\|_2^2\\
&=\left(u_i-\hat{u}_i\right)^2+\left(v_i-\hat{v}_i\right)^2
\end{aligned}
\end{equation}
where $(u_i, v_i)$ denotes the observed image coordinates and $(\hat{u}_i, \hat{v}_i)$ denotes the projected coordinates of the radar point on the image plane.

\paragraph{Optimization Objective}
As presented in Equation \ref{eq:objective}, the optimal extrinsic parameters are obtained by minimizing the sum of squared reprojection errors $\sum_{i=1}^{K} e_i(\mathbf{R}, \mathbf{t})$ across all $K$ correspondences:
\begin{equation}
\{\mathbf{R}^*, \mathbf{t}^*\} = \arg\min_{\mathbf{R}, \mathbf{t}} \sum_{i=1}^{K} e_i(\mathbf{R}, \mathbf{t})
\label{eq:objective}
\end{equation}

To avoid the orthogonality constraints on the rotation matrix, we adopt the axis-angle parameterization, converting the problem into an unconstrained six-parameter optimization. The resulting nonlinear least squares problem is solved iteratively using the Levenberg-Marquardt algorithm \cite{zhang2002flexible}, which adaptively balances between convergence speed and robustness by adjusting its damping factor.

\subsection{\underline{4D} Radar Point \underline{A}uto-\underline{L}abeling}
This section presents our radar point cloud auto-labeling method, which leverages 2D visual instance segmentation to generate point-level labels for 4D radar data. The proposed pipeline consists of three stages: 1) visual instance segmentation, which extracts object masks from RGB images using a segmentation network; 2) geometric projection coarse association, which establishes initial radar-camera correspondences by projecting 3D radar points onto the 2D image plane and matching them with instance masks; and 3) multi-feature fine optimization, which refines the initial associations by exploiting the rich information provided by 4D radar, including range, RCS and velocity, to filter false positives and recover missing points. 

\subsubsection{Problem Formulation}
Let $\mathcal{P} = \{\bm{p}_i\}_{i=1}^N$ denote a 4D radar point cloud with $N$ points. Each radar point is represented as:
\begin{equation}
\bm{p}_i = (x_i, y_i, z_i, v_i^r, \rho_i) \in \mathbb{R}^5
\end{equation}
where $(x_i, y_i, z_i)$ denotes the 3D position in radar coordinates, $v_i^r \in \mathbb{R}$ is the radial Doppler velocity, and $\rho_i \in \mathbb{R}$ is the radar RCS in dBsm.

Given a synchronized RGB image $\mathcal{I} \in \mathbb{R}^{H \times W \times 3}$, our goal is to assign each radar point $\bm{p}_i$ a label:
\begin{equation}
\ell_i = 
\begin{cases}
(c_i, id_i) & \text{if } \bm{p}_i \text{ belongs to an object} \\
\varnothing  & \text{if } \bm{p}_i \text{ is background or clutter}
\end{cases}
\end{equation}
where $c_i \in \{1, \ldots, N_c\}$ is the semantic class and $id_i \in \mathbb{Z}^+$ is the instance identifier.

\subsubsection{Visual Instance Segmentation}
We employ Mask2Former~\cite{cheng2022masked}, a state-of-the-art 2D instance segmentation method, to extract instance-level information from the RGB image. Given an input image $\mathcal{I} \in \mathbb{R}^{H \times W \times 3}$, the segmentation network $\mathcal{F}_{\text{seg}}$ with parameters $\theta_{\text{seg}}$ outputs a set of instance segmentations:
\begin{equation}
\mathcal{F}_{\text{seg}}(\mathcal{I}; \theta_{\text{seg}}) = \left\{v_j \mid v_j = \left(\mathcal{M}_j, c_j, id_j, s_j\right)\right\}_{j=1}^M
\end{equation}
where $M$ is the number of detected instances. For each instance $v_j$: $\mathcal{M}_j \in \{0, 1\}^{H \times W}$ is a binary mask that equals 1 if and only if the pixel $(u, v)$ belongs to the $j$-th instance; $c_j \in \{1, 2, \ldots, N_c\}$ denotes the category label among $N_c$ predefined classes; $id_j \in \mathbb{Z}^+$ is a unique instance identifier; and $s_j \in [0, 1]$ represents the detection confidence score, which is used for quality assessment in subsequent data association.

\subsubsection{Geometric Projection Coarse Association}
The coarse association stage establishes initial radar-camera correspondences through geometric projection. By projecting radar points onto the image plane and checking their overlap with instance segmentation masks, each point is assigned a preliminary label for subsequent refinement.

\paragraph{Point Cloud Projection}
Using the calibrated extrinsic parameters (rotation matrix $\mathbf{R}$ and translation vector $\mathbf{t}$) between the radar and camera, along with the camera intrinsic matrix $\mathbf{K}$, each radar point is projected onto the image plane. For a radar point $\bm{p}_i = (x_i, y_i, z_i)^\top$, its projected pixel coordinates $(u_i, v_i)$ are computed as:
\begin{equation}
\begin{bmatrix}
u_i \\ v_i \\ 1
\end{bmatrix}
= \frac{1}{z_i^c} \mathbf{K} \left( \mathbf{R} \begin{bmatrix} x_i \\ y_i \\ z_i \end{bmatrix} + \mathbf{t} \right)
\end{equation}
where $z_i^c$ denotes the depth of the point in the camera coordinate system, and $(u_i,v_i)$ are the projected pixel coordinates.

\paragraph{Initial Label Assignment}
After projection, each radar point is associated with the 2D instance segmentation masks by checking whether its projected pixel location falls within any mask region. For each radar point $\bm{p}_i$ with valid projection ($1 \leq u_i \leq W$, $1 \leq v_i \leq H$,  $z_i^c > 0$), we construct a candidate instance set:
\begin{equation}
\mathcal{L}_i^{\text{cand}} = \left\{ v_j \in \mathcal{V} \mid \mathcal{M}_j(u_i, v_i) = 1 \right\}
\end{equation}

Based on the candidate set, we assign a preliminary label to each radar point according to Equation~\ref{eq:coarse_label}, where points with non-empty candidate sets receive a class-instance pair $(c_{j^*}, id_{j^*})$, and these points outside all masks remain unlabeled:
\begin{equation}
\label{eq:coarse_label}
\ell_i^{\text{coarse}} = 
\begin{cases}
(c_{j^*}, id_{j^*}) & \text{if } \mathcal{L}_i^{\text{cand}} \neq \varnothing \\
\varnothing & \text{otherwise}
\end{cases}
\end{equation}

When multiple instances overlap at the projected location ($|\mathcal{L}_i^{\text{cand}}| > 1$), ambiguity arises regarding which instance the point belongs to. To address this, we select the instance with the highest confidence score, as defined in Equation~\ref{eq:max_confidence}, where $j^*$ denotes the selected instance and $s_j$ represents the confidence score of instance $v_j$:
\begin{equation}
\label{eq:max_confidence}
j^* = \arg\max_{v_j \in \mathcal{L}_i^{\text{cand}}} s_j
\end{equation}

After coarse association, we group all points sharing the same instance ID into clusters for subsequent refinement. Each cluster $\mathcal{C}_j$ is defined in Equation~\ref{eq:cluster_grouping} as the collection of points assigned to the same class-instance pair $(c_j, id_j)$. Additionally, we maintain a set $\mathcal{U}$ of unassociated points that will be considered during the point completion stage:
\begin{equation}
\label{eq:cluster_grouping}
\mathcal{C}_j = \left\{ \bm{p}_i \mid \ell_i^{\text{coarse}} = (c_j, id_j) \right\}, \quad \mathcal{U} = \{\bm{p}_i \mid \ell_i^{\text{coarse}} = \varnothing\}
\end{equation}

\subsubsection{Multi-Feature Fine Optimization}
The coarse association based on geometric projection introduces two types of errors: false positives, where points falling within a mask do not actually belong to the target (e.g., due to multi-path reflections or sensor noise), and false negatives, where points belonging to the target are missed due to occlusion, point cloud sparsity, or projection errors. To address these false positives and false negatives, we perform multi-feature fine-tuning that leverages the rich information from 4D radar to filter outliers and recover missing points.

For each point cluster $\mathcal{C}_k$ that shares the same candidate ID, we evaluate the consistency of depth, RCS, and velocity features to identify and filter outliers. We employ three complementary validation criteria to ensure robust filtering. First, for depth-based filtering, we check whether the depth $z_i^c$ of radar point $\bm{p}_i$ in the camera coordinate system is consistent with the median depth $\mu_z^j$ of cluster $\mathcal{C}_j$ within a tolerance threshold $\tau_d$, as defined in Equation~\ref{eq:depth_valid}:
\begin{equation}
\label{eq:depth_valid}
\text{DepthValid}(\bm{p}_i, \mathcal{C}_j) = 
\begin{cases}
\text{True} & \text{if } |z_i^c - \mu_z^j| < \tau_d \\
\text{False} & \text{otherwise}
\end{cases}
\end{equation}

Second, for RCS-based filtering, a point $\mathbf{p}_i \in \mathcal{C}_k$ is considered an outlier if its RCS value $\rho_i$ significantly deviates from the cluster statistics. Specifically, the deviation must not exceed $\kappa_\rho$ standard deviations from the mean, where $\mu_\rho^j$ and $\sigma_\rho^j$ denote the mean and standard deviation of RCS values in cluster $\mathcal{C}_j$, as shown in Equation~\ref{eq:rcs_valid}:
\begin{equation}
\label{eq:rcs_valid}
\text{RCSValid}(\bm{p}_i, \mathcal{C}_j) = 
\begin{cases}
\text{True} & \text{if } |\rho_i - \mu_\rho^j| \leq \kappa_\rho \cdot \sigma_\rho^j \\
\text{False} & \text{otherwise}
\end{cases}
\end{equation}

Third, for velocity-based filtering, points belonging to the same object should exhibit consistent velocity patterns. We distinguish between static and dynamic objects based on the mean radial velocity $\mu_v^j$ of the cluster. For static objects where $|\mu_v^j| \leq v_{\text{static}}$, all points are considered valid. For dynamic objects, we validate that each point's radial velocity $v_i^r$ falls within $\kappa_v$ standard deviations of the cluster mean, where $\sigma_v^j$ is the velocity standard deviation, and $\sigma_v^{\min}$ prevents over-filtering for highly consistent clusters, as expressed in Equation~\ref{eq:vel_valid}:
\begin{equation}
\label{eq:vel_valid}
\text{VelValid}(\bm{p}_i, \mathcal{C}_j) = 
\begin{cases}
\text{True} & \text{if } |\mu_v^j| \leq v_{\text{static}} \text{ (static object)} \\
\text{True} & \begin{aligned}
            &\text{if } |\mu_v^j| > v_{\text{static}} \ \land \\
            &|v_i^r - \mu_v^j| \leq \kappa_v \cdot \max(\sigma_v^j, \sigma_v^{\min})
            \end{aligned} \\[6pt]
\text{False} & \text{otherwise}
\end{cases}
\end{equation}

Overall, a point is retained in the refined cluster only if it satisfies all three validation criteria simultaneously, as expressed in Equation~\ref{eq:combined_filter}:
\begin{equation}
\label{eq:combined_filter}
\bm{p}_i \in \mathcal{C}_j^{\text{refined}} \iff \text{DepthValid} \land \text{RCSValid} \land \text{VelValid}
\end{equation}

\paragraph{In-Target Point Completion}
This step aims to recover false negative points that belong to the target but were missed during coarse association. After filtering, all unassociated radar points are obtained from a candidate set $\mathcal{U}$. For each refined cluster $\mathcal{C}_j^{\text{refined}}$, we search for neighboring points in $\mathcal{U}$ within a predefined radius $r_{\text{search}}$, as defined in Equation~\ref{eq:neighbor_search}, where $\bm{p}_u^{\text{pos}} = (x_u, y_u, z_u)^\top$ and $\bm{\mu}_{\text{pos}}^j$ is the geometric centroid of $\mathcal{C}_j^{\text{refined}}$:
\begin{equation}
\label{eq:neighbor_search}
\mathcal{U}_j^{\text{neighbor}} = \left\{ \bm{p}_u \in \mathcal{U} \mid \|\bm{p}_u^{\text{pos}} - \bm{\mu}_{\text{pos}}^j\|_2 \leq r_{\text{search}} \right\}
\end{equation}

For each candidate $\bm{p}_u \in \mathcal{U}_j^{\text{neighbor}}$, we compute a normalized affinity score based on spatial, velocity, and RCS consistency, as shown in Equation~\ref{eq:affinity}. Here, $d_{\text{pos}} = \|\bm{p}_u^{\text{pos}} - \bm{\mu}_{\text{pos}}^j\|_2$ represents the spatial distance, $d_v = |v_u^r - \mu_v^j|$ represents the velocity difference, $d_\rho = |\rho_u - \mu_\rho^j|$ represents the RCS difference, and $\sigma_{\text{pos}}$ is a spatial scale parameter:
\begin{equation}
\label{eq:affinity}
A(\bm{p}_u, \mathcal{C}_j) = \exp\left( -\frac{d_{\text{pos}}^2}{2\sigma_{\text{pos}}^2} \right) \cdot \exp\left( -\frac{d_v^2}{2(\sigma_v^j)^2} \right) \cdot \exp\left( -\frac{d_\rho^2}{2(\sigma_\rho^j)^2} \right)
\end{equation}

A candidate point is incorporated into the cluster if its affinity exceeds a threshold $\tau_A$, as expressed in Equation~\ref{eq:affinity_threshold}. When a point has high affinity with multiple clusters, we assign it to the cluster with maximum affinity according to Equation~\ref{eq:assignment}, where $\ell_u = (c_{j^*}, id_{j^*})$ denotes the assigned class and instance ID:
\begin{equation}
\label{eq:affinity_threshold}
\bm{p}_u \rightarrow \mathcal{C}_j^{\text{final}} \iff A(\bm{p}_u, \mathcal{C}_j) \geq \tau_A
\end{equation}
\begin{equation}
\label{eq:assignment}
j^* = \arg\max_j A(\bm{p}_u, \mathcal{C}_j)
\end{equation}

Finally, the complete label for each radar point is determined by Equation~\ref{eq:final_label}, where points successfully assigned to a cluster receive the cluster's class and instance ID, while unassigned points remain unlabeled:
\begin{equation}
\label{eq:final_label}
\ell_i^{\text{final}} = 
\begin{cases}
(c_j, id_j) & \text{if } \exists j: \mathbf{p}_i \in \mathcal{C}_j^{\text{final}} \\
\varnothing & \text{otherwise}
\end{cases}
\end{equation}

Overall, the complete auto-labeling pipeline is summarized in Algorithm~\ref{alg:autolabel}.

\begin{algorithm}[t]
\caption{Radar Point Cloud Auto-Labeling}
\label{alg:autolabel}
\begin{algorithmic}[1]
\REQUIRE Radar point cloud $\mathcal{P}$, RGB image $\mathcal{I}$, calibration $(\textbf{K}, \textbf{R}, \textbf{t})$
\ENSURE Point-wise labels $\{\ell_i\}_{i=1}^N$
\STATE \textbf{Stage 1: Visual Instance Segmentation}
\STATE $\mathcal{V} \gets \mathcal{F}_{\text{seg}}(\mathcal{I})$ {Extract instances}
\STATE
\STATE \textbf{Stage 2: Geometric Projection Coarse Association}
\FOR{each point $\bm{p}_i \in \mathcal{P}$}
    \STATE $(u_i, v_i) \gets \pi(\bm{p}_i)$ {Project to image}
    \STATE $\mathcal{L}_i^{\text{cand}} \gets \{v_j \mid \mathcal{M}_j(u_i, v_i) = 1\}$
    \IF{$\mathcal{L}_i^{\text{cand}} \neq \varnothing$}
        \STATE $j^* \gets \arg\max_{v_j \in \mathcal{L}_i^{\text{cand}}} s_j$
        \STATE $\ell_i^{\text{coarse}} \gets (c_{j^*}, id_{j^*})$
    \ELSE
        \STATE $\ell_i^{\text{coarse}} \gets \varnothing$
    \ENDIF
\ENDFOR
\STATE Form clusters $\{\mathcal{C}_j\}$ and unassociated set $\mathcal{U}$
\STATE
\STATE \textbf{Stage 3: Multi-Feature Fine OptimizationRefinement}
\FOR{each cluster $\mathcal{C}_j$ with $|\mathcal{C}_j| \geq n_{\min}$}
    \STATE Compute statistics $(\mu_z^j, \mu_\rho^j, \sigma_\rho^j, \mu_v^j, \sigma_v^j)$
    \STATE \textbf{Filtering:} Remove points violating depth/RCS/velocity criteria
    \STATE \textbf{Recovery:} Add nearby points from $\mathcal{U}$ with $A(\bm{p}_u, \mathcal{C}_j) \geq \tau_A$
\ENDFOR
\STATE Assign final labels $\{\ell_i^{\text{final}}\}$
\end{algorithmic}
\end{algorithm}

\section{Experiments}
\label{sec:experiments}
This section presents comprehensive experiments to evaluate the proposed 4D-CAAL method. We first describe the experimental setup, including hardware configuration, data collection protocol, and evaluation metrics. Then, we report quantitative results on calibration accuracy, followed by ablation studies to analyze the contribution of key components. Finally, we provide qualitative visualizations to demonstrate the effectiveness of our method.

\subsection{Experimental Setup}

\subsubsection{Hardware Configuration}
Our sensor suite consists of a 4D imaging radar and a monocular RGB camera, rigidly mounted on the custom bracket described in Figure \ref{fig:fixed}. The detailed sensor specifications are listed in Table \ref{tab:sensor-specs}.

\begin{table}[t]
\centering
\caption{Sensor specifications used in our experiments.}
\label{tab:sensor-specs}
\begin{tabular*}{\linewidth}{
p{3cm}<{}|
p{2.3cm}<{\centering}
p{2.3cm}<{\centering}
}
\toprule
\textbf{Parameter} & \textbf{4D Radar} & \textbf{Camera} \\
\midrule
Model & Oculii EAGLE & SONY IMX317 \\
Range & 0.5--200 m & -- \\
Field of View (H $\times$ V) & 110$^\circ$ $\times$ 45$^\circ$  & 100$^\circ$ $\times$ 60$^\circ$ \\
Angular Resolution & 0.5$^\circ$ $\times$ 1.0$^\circ$ & -- \\
Resolution & -- & 1920 $\times$ 1080 \\
Frame Rate & 15 Hz & 30 Hz \\
\bottomrule
\end{tabular*}
\end{table}

\subsubsection{Implementation Details}
To evaluate the proposed calibration procedure, an outdoor experiment was conducted. The calibration target was placed at various positions and orientations within the overlapping field of view of both sensors. In total, $K = 24$ calibration poses were collected, ensuring diverse spatial coverage. For each pose, synchronized radar point clouds and camera images were recorded.

The proposed calibration pipeline was implemented in Python, utilizing OpenCV for camera feature extraction and Open3D for radar point cloud processing. Levenberg-Marquardt optimization was performed using the SciPy library. The key parameters used in our experiments are summarized in Table~\ref{tab:calibration-params}.

\begin{table}[t]
\centering
\caption{Parameter settings for the calibration pipeline.}
\label{tab:calibration-params}
\begin{tabular*}{\linewidth}{
p{1.2cm}<{}|
p{1.1cm}<{\centering}
p{3.6cm}<{\centering}
p{1.3cm}<{\centering}
}
\toprule
\textbf{Category} & \textbf{Parameter} & \textbf{Description} & \textbf{Value} \\
\midrule
\multirow{4}{*}{Filtering} 
& $R_{\min}$ & Minimum detection range & 3.0 m \\
& $R_{\max}$ & Maximum detection range & 15.0 m \\
& $v_{\text{th}}$ & Static velocity threshold & 0.5 m/s \\
& $\rho_{\min}$ & Minimum RCS threshold & 10.0 dBsm \\
\midrule
\multirow{2}{*}{Clustering} 
& $\epsilon$ & DBSCAN neighborhood radius & 0.3 m \\
& $N_{\min}$ & DBSCAN minimum points & 3 \\
\bottomrule
\end{tabular*}
\end{table}

For radar point auto-labeling, we employ Mask2Former~\cite{cheng2022masked} with a Swin-L backbone pretrained on the COCO dataset and fine-tuned on the nuScenes dataset. The model outputs instance masks at an input resolution of $1920 \times 1080$ pixels.

Radar points are projected onto the image plane using pre-calibrated extrinsic and intrinsic parameters. Points falling outside the camera field of view or behind the camera plane are excluded from association. The key parameters for the auto-labeling pipeline are summarized in Table~\ref{tab:labeling-params}.

\begin{table}[t]
\centering
\caption{Parameter settings for the auto-labeling stage.}
\label{tab:labeling-params}
\begin{tabular*}{\linewidth}{
p{1.4cm}<{}|
p{1.1cm}<{\centering}
p{3.4cm}<{\centering}
p{1.3cm}<{\centering}
}
\toprule
\textbf{Category} & \textbf{Parameter} & \textbf{Description} & \textbf{Value} \\
\midrule
\multirow{6}{*}{\makecell[l]{Outlier\\Filtering}} 
& $\tau_d$ & Depth tolerance & 1.5 m \\
& $\kappa_\rho$ & RCS deviation multiplier & 2.5 \\
& $\kappa_v$ & Velocity deviation multiplier & 2.0 \\
& $v_{\text{static}}$ & Static object threshold & 0.3 m/s \\
& $\sigma_v^{\min}$ & Minimum velocity std. & 0.2 m/s \\
\midrule
\multirow{3}{*}{\makecell[l]{Point\\Completion}} 
& $r_{\text{search}}$ & Point completion radius & 2.0 m \\
& $\sigma_{\text{pos}}$ & Spatial scale parameter & 0.8 m \\
& $\tau_A$ & Affinity threshold & 0.6 \\
\bottomrule
\end{tabular*}
\end{table}

\subsubsection{Evaluation Metrics}
We evaluate the proposed framework using two categories of metrics: reprojection error metrics for assessing calibration accuracy and matching accuracy metrics for evaluating annotation quality. These complementary metrics provide a comprehensive assessment of both components in our 4D-CAAL framework.

\paragraph{Reprojection Error Metrics}
Reprojection error quantifies the discrepancy between observed 2D image points and their corresponding 3D radar points projected onto the image plane using the estimated extrinsic parameters. This metric directly reflects calibration quality in the image domain, which is most relevant to downstream sensor-fusion applications.

We employ Mean Reprojection Error (MRE) and Root Mean Square Error (RMSE) to assess calibration accuracy. MRE measures the average Euclidean distance between observed and projected image points across all calibration correspondences:
\begin{equation}
\begin{aligned}
\text{MRE} &= \frac{1}{K} \sum_{i=1}^{K} \|\mathbf{c}_i - \hat{\mathbf{c}}_i\|_2 \\
&= \frac{1}{K} \sum_{i=1}^{K} \sqrt{(u_i - \hat{u}_i)^2 + (v_i - \hat{v}_i)^2}
\end{aligned}
\end{equation}
where $\mathbf{c}_i = (u_i, v_i)^\top$ denotes the observed image coordinates, $\hat{\mathbf{c}}_i = (\hat{u}_i, \hat{v}_i)^\top$ denotes the projected coordinates, and $K$ is the total number of calibration poses. MRE provides an intuitive measure of average calibration accuracy in pixel units, with lower values indicating better alignment between the two sensor modalities.

RMSE computes the root mean square of reprojection errors, penalizing larger errors more heavily than MRE:
\begin{equation}
\begin{aligned}
\text{RMSE} &= \sqrt{\frac{1}{K} \sum_{i=1}^{K} \|\mathbf{c}_i - \hat{\mathbf{c}}_i\|_2^2} \\
&= \sqrt{\frac{1}{K} \sum_{i=1}^{K} \left[(u_i - \hat{u}_i)^2 + (v_i - \hat{v}_i)^2\right]}
\end{aligned}
\end{equation}
Compared to MRE, RMSE is more sensitive to outliers, making it particularly useful for detecting calibration inconsistencies and identifying potential failure cases.

\paragraph{Matching Accuracy Metrics}
Matching accuracy evaluates the quality of point-to-instance associations in the automatic annotation pipeline. These metrics assess how accurately radar points are assigned to their corresponding object instances compared to ground truth annotations.

We employ Point Accuracy (PA) and Mean Intersection over Union (mIoU) to evaluate annotation quality. PA measures the percentage of points correctly assigned to their corresponding instances:
\begin{equation}
\text{PA} = \frac{N_{\text{correct}}}{N_{\text{total}}}
\end{equation}
where $N_{\text{correct}}$ denotes the number of correctly labeled points and $N_{\text{total}}$ denotes the total number of annotated points. PA provides a straightforward measure of overall annotation correctness at the point level.

mIoU computes the average Intersection over Union across all matched instance pairs, evaluating the overlap between predicted and ground truth point sets:
\begin{equation}
\text{mIoU} = \frac{1}{N_{\text{matched}}} \sum_{i=1}^{N_{\text{matched}}} \text{IoU}_i, \quad \text{IoU}_i = \frac{|\mathcal{P}_{\text{pred}}^i \cap \mathcal{P}_{\text{gt}}^i|}{|\mathcal{P}_{\text{pred}}^i \cup \mathcal{P}_{\text{gt}}^i|}
\end{equation}
where $\mathcal{P}_{\text{pred}}^i$ and $\mathcal{P}_{\text{gt}}^i$ denote the predicted and ground truth point sets for instance $i$, respectively, and $N_{\text{matched}}$ is the number of matched instances. mIoU captures both precision and recall aspects of annotation quality, providing a balanced assessment of instance-level labeling performance.

\subsection{Experimental Results}
\subsubsection{Calibration Evaluation}
Table \ref{tab:calibration-results} summarizes the calibration accuracy in the test pairs from data collection. The results demonstrate that our approach achieves higher reprojection accuracy, with an MRE value of 5.25 pixels and an RMSE value of 8.76 pixels. It is noteworthy that currently there is no publicly available 4D radar-camera dataset specifically designed for calibration purposes. The result reported for the baseline method was obtained using a TI AWR1843BOOST radar with a targetless strategy.

\begin{table}[t]
\centering
\caption{Comparison of calibration accuracy with baseline methods.}
\label{tab:calibration-results}
\begin{tabular*}{\linewidth}{
p{2.3cm}<{}|
p{2.8cm}<{\centering}
p{2.2cm}<{\centering}
}
\toprule
\makecell[l]{\textbf{Method}} &  \makecell{Cheng \etal  \cite{cheng2023online}} &  \makecell{\textbf{4D-CAAL}  (ours)}  \\
\midrule
\makecell[l]{\textbf{Radar Type}} & \makecell{4D \\ (TI AWR1843BOOST)} & \makecell{4D \\ (Oculii EAGLE)} \\
\midrule
\makecell[l]{\textbf{Target Type}} & Targetless & \makecell{Corner Reflector \\ \& Checkerboard} \\
\midrule
\makecell[l]{\textbf{MRE} (pixels) $\downarrow$} & 59.89 & \textbf{5.25} \\
\midrule
\makecell[l]{\textbf{RMSE} (pixels) $\downarrow$} & 98.48 & \textbf{8.76} \\
\bottomrule
\end{tabular*}
\end{table}

\subsubsection{Auto-Labeling Evaluation}
\begin{table}[t]
\centering
\caption{Comparison of auto-labeling quality with baseline methods. OTPE: Out-of-Target Point Filtering. ITPC: In-Target Point Completion.}
\label{tab:auto-labeling-results}
\begin{tabular*}{\linewidth}{
p{2.8cm}<{}|
p{2.4cm}<{\centering}
p{2.4cm}<{\centering}
}
\toprule
\textbf{Configuration} & \textbf{PA (\%)} $\uparrow$ & \textbf{mIoU (\%)} $\uparrow$ \\
\midrule
A0: Baseline & 85.73 & 72.41 \\\midrule
A1: \quad + Depth & 86.91 (+1.18) & 73.82 (+1.41)  \\
A2: \quad + RCS & 86.14 (+0.41) & 72.93 (+0.52)\\
A3: \quad + Velocity & 86.58 (+0.85) & 73.47 (+1.06) \\
A4: \quad + All OTPF & 88.26 (+2.53) & 75.38 (+2.97)  \\
A5: \quad + All ITPC & 86.21 (+0.48) & 74.12 (+1.71) \\\midrule
\textbf{4D-CAAL} (ours) & \textbf{90.12} (+4.39) & \textbf{77.83} (+5.42) \\
\bottomrule
\end{tabular*}
\end{table}

Table~\ref{tab:auto-labeling-results} presents the ablation study results for our auto-labeling pipeline. The baseline (A0) employs only geometric projection-based coarse association, achieving 85.73\% PA and 72.41\% mIoU. 
Among the individual filtering features, depth filtering (A1) provides the largest improvement (+1.18\% PA, +1.41\% mIoU), as depth consistency effectively distinguishes foreground objects from background clutter. Velocity filtering (A3) yields moderate gains (+0.85\% PA, +1.06\% mIoU), beneficial for dynamic objects with consistent radial velocities. RCS filtering (A2) contributes the smallest improvement (+0.41\% PA, +0.52\% mIoU), due to the inherent variability of RCS values across different parts of the same object.

The combined OTPF module (A4) achieves +2.53\% PA and +2.97\% mIoU over the baseline, with gains less than the sum of individual contributions due to partial overlap. The ITPC module (A5) shows a modest improvement in PA (+0.48\%) but a more substantial gain in mIoU (+1.71\%), confirming that affinity-based point recovery primarily improves recall.
The complete 4D-CAAL framework achieves 90.12\% PA and 77.83\% mIoU, with improvements of +4.39\% and +5.42\% respectively. The results demonstrate that OTPF and ITPC are complementary: OTPF improves precision by removing spurious associations, while ITPC enhances recall by recovering missed points.

\subsection{Discussion}
For calibration, we design a joint target that combines a corner reflector with a checkerboard. This design leverages the strong radar response of the corner reflector and the precise corner detection capability of the checkerboard, enabling robust cross-modal feature association. Consequently, it overcomes the limitations of targetless methods in scenarios with sparse features and high noise sensitivity.

For auto-labeling, we propose a coarse-to-fine annotation strategy that combines visual semantic understanding with 4D radar-specific characteristics. As demonstrated in our ablation study (Table \ref{tab:auto-labeling-results}), the OTPF module improves precision by filtering false positives near object boundaries, while the ITPC module improves recall by recovering missed points due to occlusion and sensor sparsity. These results validate the effectiveness of our approach in generating high-quality point-level labels, making 4D-CAAL a practical and scalable solution for large-scale 4D radar dataset construction.

Despite achieving accurate results, our method has several limitations. For calibration, the accuracy depends on the manufacturing precision of the corner reflector and its alignment with the center of the checkerboard. Manual placement in multiple poses can also be time-consuming for large-scale deployment. For auto-labeling, annotation quality is bounded by visual instance segmentation performance, which may degrade under severe occlusion or extreme lighting. Additionally, significant object overlap in the image plane leads to ambiguous point-to-instance correspondences.

Future work will focus on three directions. First, we plan to develop online calibration methods to handle sensor drift during vehicle operation. Second, we aim to introduce temporal consistency constraints across consecutive frames to enhance the robustness of the annotation. Third, we will explore self-supervised or weakly supervised approaches to reduce reliance on pretrained segmentation models and improve generalization across diverse scenarios.

\section{Conclusion}
\label{sec:conclusion}
In this study, we presented 4D-CAAL, a comprehensive framework for 4D radar-camera calibration and point cloud auto-labeling. For extrinsic calibration, we designed a hybrid calibration target integrating a trihedral corner reflector with a checkerboard pattern, combined with systematic feature extraction and nonlinear optimization, achieving high reprojection accuracy with robust convergence. For point cloud auto-labeling, we proposed a coarse-to-fine strategy that establishes initial radar-camera correspondences through geometric projection and refines them by leveraging 4D radar-specific features, including RCS and velocity. Experimental results demonstrate that our 4D-CAAL framework provides accurate extrinsic calibration and high-quality automatic annotations comparable to manual labels.

With the rapid advancement of radar-camera fusion in autonomous driving, scalable dataset construction has become increasingly critical. The proposed 4D-CAAL framework addresses this need by significantly reducing the manual effort required for sensor calibration and data annotation, thereby lowering the barrier to developing and deploying 4D radar-based perception systems. Future work will focus on online calibration adaptation to handle sensor drift during vehicle operation, temporal consistency across consecutive frames to improve annotation robustness, and self-supervised approaches to reduce reliance on pretrained models and enhance generalization across diverse driving scenarios.

\bibliographystyle{IEEEtran}
\bibliography{citations,others}

\end{document}